\pdfoutput=1

\documentclass[11pt]{article}

\usepackage[]{EACL2023}

\usepackage{times}
\usepackage{latexsym}

\usepackage[T1]{fontenc}

\usepackage{microtype}

\usepackage{inconsolata}

\usepackage{paralist}

\usepackage[utf8]{inputenc}

\usepackage{amsmath}
\usepackage{graphicx}

\title{Weakly-Supervised Questions for Zero-Shot Relation Extraction}

\author{Saeed Najafi \and
  Alona Fyshe \\
  Department of Computing Science, University of Alberta, Canada\\
  \texttt{\{snajafi,alona\}@ualberta.ca} \\}
  
\begin{document}
\maketitle
\begin{abstract}
Zero-Shot Relation Extraction (ZRE) is the task of Relation Extraction where the training and test sets have no shared relation types. This very challenging domain is a good test of a model's ability to generalize. Previous approaches to ZRE reframed relation extraction as Question Answering (QA), allowing for the use of pre-trained QA models. However, this method required manually creating gold question templates for each new relation. Here, we do away with these gold templates and instead learn a model that can generate questions for unseen relations. Our technique can successfully translate relation descriptions into relevant questions, which are then leveraged to generate the correct tail entity.  On tail entity extraction, we outperform the previous state-of-the-art by more than 16 F1 points without using gold question templates.  On the RE-QA dataset where no previous baseline for relation extraction exists, our proposed algorithm comes within 0.7 F1 points of a system that uses gold question templates. Our model also outperforms the state-of-the-art ZRE baselines on the FewRel and WikiZSL datasets, showing that QA models no longer need template questions to match the performance of models specifically tailored to the ZRE task. Our implementation is available at \url{https://github.com/fyshelab/QA-ZRE}.
\end{abstract}

\section{Introduction}
Building models that capture abstract knowledge rather than just memorizing data is one of the inspirations for zero-shot benchmarks \cite{DBLP:journals/corr/abs-2109-14076}. In order to test a model's ability to extract higher-level knowledge, these benchmarks measure performance on data unlike what is seen during training.
For example, in Zero-shot Relation Extraction (ZRE), a model trained to extract the name of a company's CEO should be able to extract the name of a country's political leader at test time.

In ZRE, the test relations do not appear in the training data, so one cannot apply typical relation classification approaches. One method for ZRE is to reframe the task as a Question-Answering (QA) problem by manually creating question templates for each relation type. Extracting the tail entity is accomplished by finding the answer span for the corresponding question template~\cite{levy-etal-2017-zero}. However, this method requires gold question templates ahead of time. Here we ask, is it possible to leverage QA systems without annotating relations with gold question templates?

We treat ZRE as a tail entity generation task for which we consider the dual training of question and answer generators. We create question and answer generators by pre-training the publicly available T5 models \cite{JMLR:v21:20-074} on QA corpora, and then fine-tune for ZRE within our proposed learning framework. We investigate four training objectives based on Marginal Maximum Likelihood (MML) optimization, and suggest an off-policy sampling technique to avoid ungrammatical spurious questions in the search space. 

Our experiments show that our off-policy sampling technique is critical for creating semantically relevant questions. For ZRE, we show that our weakly-supervised questions produce a model with competitive F1 score of 65.4 compared to the F1 score of 66.1 achieved by using gold question templates. Our contributions can be summarized as the following:

\begin{compactitem}
     \item We propose a new learning objective that combines off-policy sampling and MML optimization.
     \item We can successfully generate semantically relevant questions for a given relationship signal.
     \item We report a new state-of-the-art ZRE performance on the RE-QA dataset~\cite{levy-etal-2017-zero} without using gold question templates.
     \item We outperform SOTA baselines on the FewRel and WikiZSL datasets with a generative approach without using any gold question templates.
 \end{compactitem}

\section{Problem Formulation}
In this work, we train models to extract facts from unstructured text.  Facts are represented as triplets $(e_1, r, e_2)$ where $e_1$ is the head entity, $e_2$ is the tail entity, and $r$ is the relation keywords. We explore \textbf{relation extraction (RE)}, which is the task of predicting $r$ when $e_1$ and $e_2$ are known, and \textbf{tail entity extraction (TE)}, which is the task of predicting $e_2$ when $e_1$ and $r$ are known.

We begin with TE, for which we learn $P(e_2 |c, e_1, r)$: the distribution of the tail entity conditioned on an unstructured text context $c$, head entity, and the relation keywords. We investigate how to transfer models pre-trained on QA corpora to the task of TE by estimating $P(e2|c, e1, r)$. With a gold natural question $q^{*}$, which semantically includes the information about both of the head entity $e_1$ and the relation words $r$, we can optimize $P(e2|c, q^{*})$ to fine-tune any answer generation model on RE datasets\footnote{On an RE dataset, we can fine-tune models for TE and ZRE tasks.}. In cases where we do not have gold question templates, we can define a simple baseline using a pseudo question $q^{pseudo}$, which is the string concatenation of
the head entity $e_1$ and the relation words $r$. We then can fine-tune a pre-trained QA model on RE datasets by directly optimizing the objective $P(e2|c, q^{pseudo})$.

\begin{figure}[h]
\begin{center}
\includegraphics[width=0.4\textwidth]{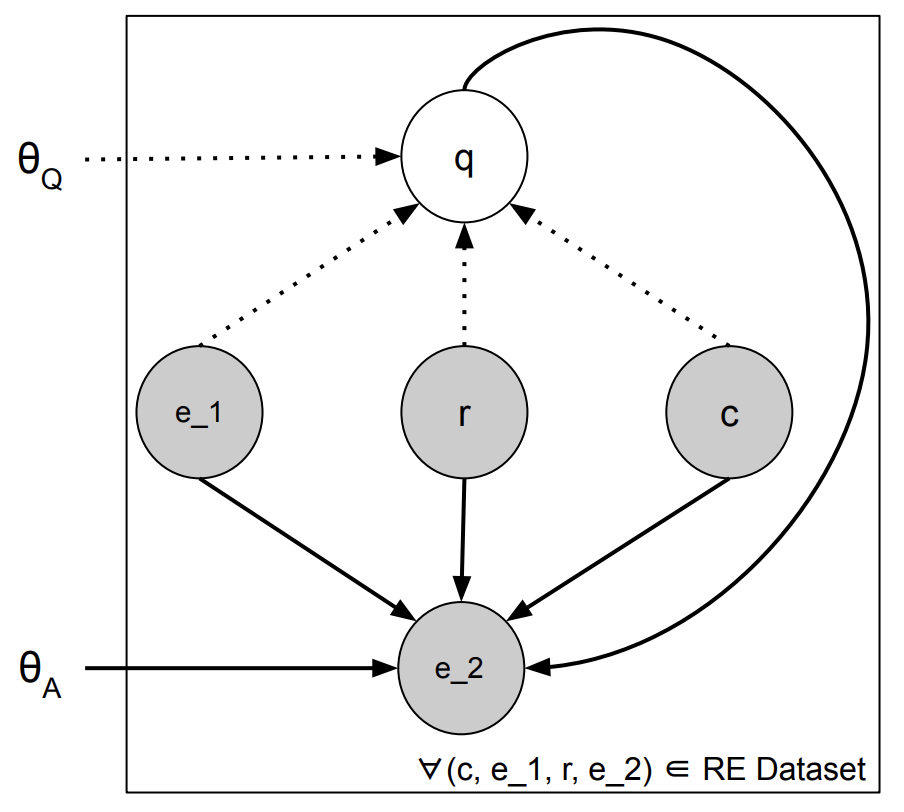}
\end{center}
\caption{
The probabilistic representation of the conditional distribution of the tail entity ($e_2$) conditioned on the context $c$, head entity $e_1$, and the relation keywords
$r$. The question $q$ is not observed. The $\theta_Q$ and $\theta_A$ are respectively the parameters of the question and answer generation modules.}
\label{pgm-figure}
\end{figure}

To transfer the models pre-trained on QA corpora to the task of TE, we can provide a natural question for the given head entity and the relation keywords. However, providing question templates for every relation type is infeasible. We explore the idea of generating questions semantically relevant to the given input context, relation and the head entity. Therefore, we marginalize the joint distribution $P(e_2, q | c, e_1, r)$ with respect to the unobserved questions $q$ to learn the tail entity generator: $P(e_2 | c, e_1, r) = \sum_{q} P(e_2, q | c, e_1, r)$. As illustrated by Figure \ref{pgm-figure}, we factorize $P(e_2, q| c, e_1, r)$ as the following equation where $\theta_Q$ and $\theta_A$ are respectively the parameters of the question and answer generation modules:
\begin{multline}
    P(e_2 | c, e_1, r) = \sum_{q} P(e_2, q | c, e_1, r) = \\ \sum_{q} P_{\theta_Q} (q | c, e_1, r) \times P_{\theta_A} (e_2 | c , q, e_1, r)
\label{main-tail-objective}
\end{multline}

For example, given a context biography about the person ``Donald Trump'' and the relation keywords ``place of birth,'' we would use $P_{\theta_Q}$ to first generate the question ``Where was Donald Trump born?'' and then the answer module $P_{\theta_A}$ can generate the response ``New York.'' 

We then re-purpose this TE model for the task of RE.  Whether we train the tail entity generator $P(e_2 | c, e_1, r)$ with gold, pseudo, or generated questions, we perform the RE task by scoring every possible relation of the test data and choosing the highest scoring relation:
\begin{equation}
    \hat{r} = \arg\max_{r} P(e_2 | c, e_1, r)
\label{re-class-inference}
\end{equation}

\section{Methods}
In this section we discuss the pre-training steps
for the question and answer modules and then
explore combinations of four training objectives based on maximizing the marginal likelihood of Equation \ref{main-tail-objective}. Finally, we introduce off-policy sampling and summarize the search algorithms used during training and test phases.

\subsection{Pre-training the Answer Module}
The answer generator is based on the publicly available T5 model~\cite{JMLR:v21:20-074}. We fine-tune this T5 model on the following five QA datasets to generate the gold answer in its output given the input passage and the question: 1) SQuAD~\cite{rajpurkar-etal-2018-know} with unanswerable questions by generating a special unknown token in the output corresponding to these unanswerable questions, 2) NarrativeQA dataset containing abstractive responses for its questions~\cite{kocisky-etal-2018-narrativeqa}, 3) RACE dataset as a multiple-choice QA dataset where we generate the question's correct choice as the answer text in the decoder~\cite{lai-etal-2017-race}, 4) BoolQ dataset having Yes/No questions~\cite{clark-etal-2019-boolq}, and 5) the DropQ dataset to encourage the model to learn discrete reasoning (e.g., addition, counting or sorting, etc.) in paragraphs~\cite{dua-etal-2019-drop}. We fine-tune the T5 on the aforementioned QA corpora for four epochs with a batch size of 64. In total, we have fine-tuned this T5 model on 337,768 distinct QA pairs. We use the final model as our pre-trained answer generator which will be the initial checkpoint for learning the distributions $P(e_2 | c, q^{*})$, $P(e_2 | c, q^{pseudo})$, or $P_{\theta_A} (e_2 | c , q, e_1, r)$ in our experiments.

\subsection{Pre-training the Question Module}
We now fine-tune a T5 model using the QA corpora.  We learn the distribution $P (q | c, e_1, r, e_2)$, and use the answer in each QA instance as a proxy for the tail entity $e_2$. To specify the synthetic head entity $e_1$, we run a named entity tagger\footnote{We used the NER tagger from the SpaCy library.} over questions in the QA corpus. For questions with multiple extracted entities, one of the entities are randomly selected as the head entity. To specify the relation $r$ from the question, after ignoring punctuation or interrogative words (e.g., what, where, etc.), at most four tokens (excluding the extracted entities) are sampled. We fine-tune the T5 model on this synthetic dataset, giving gold passage as the context $c$, one of the extracted named entities as $e_1$, the gold answer tokens as $e_2$, and the sampled words as the relationship keyword $r$. We train the model to produce the gold question. We specifically fine-tune the model on the answerable questions from the SQuAD~\cite{rajpurkar-etal-2018-know}, NarrativeQA~\cite{kocisky-etal-2018-narrativeqa}, and the DropQ datasets~\cite{dua-etal-2019-drop} for 4 epochs. This model will be the initial checkpoint for the question generator $P_{\theta_Q} (q | c, e_1, r)$ in our experiments.

\subsection{Training Objectives}
Having the pre-trained answer generator, we optimize $\log P(e_2 | c, q^{*})$, and $\log P(e_2 | c, q^{pseudo})$ on RE datasets using the gold $q^*$ and pseudo questions $q^{pseudo}$, respectively. To generate questions and learn from them, we also directly maximize the Marginal Log-Likelihood (MML) to train both the question and answer modules in Equation \ref{main-tail-objective}. With MML training, the objective is approximated with numerical summation using questions sampled from the question module. MML training outperforms the policy gradient methods on  weakly-supervised semantic parsing \cite{guu-etal-2017-language} as simple policy gradient estimations may have large variances \cite{NIPS2008_8df707a9, guu-etal-2017-language}.
With the MML training, we use the following gradient vector to update $\theta_Q$, where $\phi(q)$ approximates the true question posterior $P (q | c, e_1, r, e_2)$:
\begin{multline}
\nabla_{\theta_Q} M = \sum_{q} \phi(q) \times \nabla_{\theta_Q} \log P_{\theta_Q} (q | c, e_1, r)\\
\phi(q) = \frac{P_{\theta_Q} (q | c, e_1, r) \times P_{\theta_A} (e_2 | c , q, e_1, r)}{Z} \\
Z = \sum_{q^{'}} P_{\theta_Q} (q^{'} | c, e_1, r) \times P_{\theta_A} (e_2 | c , q^{'}, e_1, r)
\label{q-mml-objective}
\end{multline}

Similarly, we use the following gradient update for $\theta_A$: $\nabla_{\theta_A} M =$
\begin{multline}
\sum_{q} \phi(q) \times \nabla_{\theta_A} \log P_{\theta_A} (e_2 | c , q, e_1, r)
\label{a-mml-objective}
\end{multline}

As we ultimately want to generate the correct tail entity regardless of the input question in the answer module, we can use the best question from $P_{\theta_Q}$ and then optimize the probability of the tail entity for such a decoded question. This idea results in the following G gradient (G: greedy) update for the answer module:
\begin{multline}
\nabla_{\theta_A} G =
     \nabla_{\theta_A} \log P_{\theta_A} (e_2 | c , \hat{q}, e_1, r) \\ 
     \hat{q} = \arg\max_{q} P_{\theta_Q} (q | c, e_1, r)
\label{answer-pg-g-gradient}
\end{multline}

\subsection{Off-Policy Search over Pre-trained Question Posterior}
We encountered two issues while training the question generator with the previous MML objective. 
The first issue is that samples become ungrammatical templates as we continue training the question module. We hypothesize that the issue originates from taking direct samples from the question generator $P_{\theta_Q} (q | c, e_1, r)$ while simultaneously updating it during training. These spurious questions can hinder the model from generalizing to a new unseen relation. The program synthesis and semantic parsing research have reported a similar issue~\cite{guu-etal-2017-language, misra-etal-2018-policy, NEURIPS2018_f4e369c0, wang-etal-2019-learning, wang-etal-2021-learning}. 

The second issue in taking samples from $P_{\theta_Q} (q | c, e_1, r)$ is that omitting the tail entity in the question generator while using the input context $c$ sometimes results in multiple plausible questions. For example, consider the context ``The Facebook brand was replaced by Meta in November 2021.'' For the ambiguous relationship word ``replace'', given the head entity ``Facebook'', we can generate two valid questions: a) ``Which brand replaced Facebook?", b) ``When was the brand Facebook replaced?".

To resolve the first issue, we apply an off-policy sampling technique~\cite{DBLP:journals/corr/abs-2005-01643, DBLP:journals/corr/abs-2009-07839} and use a fixed separate sampling module to sample questions during training. With this fixed search policy $S(q)$ over questions, we need to optimize the following new objective:
\begin{multline*}
    \log E_{q \sim S(q)} [\frac{P_{\theta_Q} (q | c, e_1, r)}{S(q)} \times P_{\theta_A} (e_2 | c , q, e_1, r)]
\end{multline*}

To resolve the second issue, we feed the tail entity along with the context into the search module; however, we cannot feed the tail entity into the question generator $P_{\theta_Q} (q | c, e_1, r)$ as the correct tail entity is unknown during the test phase. Furthermore, to augment the information of the relationship keyword $r$, we append the relation description for each $r$ and feed it to the question generator $P_{\theta_Q}$ and the sampling module $S(q)$.

We assume that our search module $S(q)$ is a fixed model approximating the question posterior $P(q | c, e_1, r, e_2)$, which can be our pre-trained question generator. Thus, we use two identical copies of our pre-trained question generator. One is fixed and will be used to provide samples for training. The other serves as the initial network for $P_{\theta_Q} (q | c, e_1, r)$, and we update its parameters during the training phase. The search module will also receive the tail entity $e_2$ as input, whereas the $P_{\theta_Q}$ only receives $c$, $e_1$, and $r$. During the test phase, we only generate questions from $P_{\theta_Q}$. Using the off-policy samples, the new MML gradient updates will have the following forms, where $S(q)$ is our pre-trained question module and $\phi(q)$ approximates the true question posterior according to Equation \ref{q-mml-objective}:
\begin{align}
\nabla_{\theta_Q} O &= E_{q \sim S(q)} [\frac{\phi(q)}{S(q)} \times \nabla_{\theta_Q} \log P_{\theta_Q} (q | c, e_1, r)] 
\label{OffMML-objective-Q}\\
\nabla_{\theta_A} O &= E_{q \sim S(q)} [\frac{\phi(q)}{S(q)} \times \nonumber\\
&\nabla_{\theta_A} \log P_{\theta_A} (e_2 | c, q, e_1, r)]
\label{OffMML-objective-A}
\end{align}

Table \ref{objectives} summarizes the four gradient vectors we use for the training phase. With $\nabla_{\theta_Q} G$ gradient update, we always use samples from the question generator (not from $S(q)$) to expose the answer module to errors from $P_{\theta_Q}$.

\begin{table}[tb]
\centering
\begin{tabular}{l | ll}
\hline
\textbf{Training Objective} & Q grad. & A grad.\\
\hline
MML-MML & $\nabla_{\theta_Q} M$ & $\nabla_{\theta_A} M$\\
MML-G & $\nabla_{\theta_Q} M$ & $\nabla_{\theta_A} G$\\
OffMML-OffMML & $\nabla_{\theta_Q} O$ & $\nabla_{\theta_A} O$\\
OffMML-G & $\nabla_{\theta_Q} O$ & $\nabla_{\theta_A} G$\\
\hline
\end{tabular}
\caption{\label{objectives}
Combinations of the four types of gradient vectors used for training the question ($\theta_Q$) and answer ($\theta_A$) modules.  See Equations~\ref{q-mml-objective}-\ref{OffMML-objective-A}. Training objectives are named as (Question-Answer) tuples that refer to the gradient vectors used for the question and answer modules.
}
\end{table}

\subsection{Search Algorithms}
To generate the tail entity on the test data, regardless of fine-tuning the pre-trained answer module with gold, pseudo, or generated questions, we use greedy decoding to find the top-scoring tokens. To generate questions from $P_{\theta_Q} (q | c, e_1, r)$, we use the top-scoring question found with beam search decoding having the beam size of 8.

During training, to estimate the $G$ gradient vector listed in Table \ref{objectives}, we rely on beam search decoding with the beam size of 8 to find the top scoring question. For MML gradient vectors, we use top-p sampling with the suggested threshold of $0.95$ to collect $8$ final samples from the question generator $P_{\theta_Q} (q | c, e_1, r)$ or sampling module $S(q)$. In top-p sampling \cite{holtzman2019curious}, at each decoding step, we find the smallest set of vocabulary with the cumulative probability above a threshold (e.g., $0.95$) and re-scale the distribution among these tokens to perform the sampling. We find this approach has a better exploration compared to beam-search, greedy, or top-k decoding.

\section{Experiments Setup}

\begin{table*}
\centering
\caption{The example input for the question and search modules. The search module also receives the tail entity in the input.}
\begin{tabular}{ p{0.08\linewidth} | p{0.84\linewidth} }
\hline
Module & Input Format\\
\hline
Question & ``answer: Isaac Nicola <SEP> place of birth ; most specific known birth location of a person, animal or fictional character context: Isaac Nicola Romero (1916 in Havana, Cuba) was a prominent Cuban guitarist. </s>''\\
\hline
Search & ``answer: Isaac Nicola <SEP> place of birth ; most specific known birth location of a person, animal or fictional character Havana context: Isaac Nicola Romero (1916 in Havana, Cuba) was a prominent Cuban guitarist. </s>'' \\
\hline
\end{tabular}
\label{q_input_formats}
\end{table*}

\begin{table*}
\centering
\caption{The example input for the answer module given gold, pseudo, and the generated questions. The generated question is from the question module trained with the OffMML-G objective.}
\begin{tabular}{ p{0.14\linewidth} | p{0.78\linewidth} }
\hline
Question Type & Input Format\\
\hline
Pseudo & ``question: Isaac Nicola <SEP> place of birth context: Isaac Nicola Romero (1916 in Havana, Cuba) was a prominent Cuban guitarist. </s>''\\
\hline
Generated & ``relation:  Isaac Nicola  place of birth question: What was the name of the place where Isaac Nicola was born? context: Isaac Nicola Romero (1916 in Havana, Cuba) was a prominent Cuban guitarist. </s>'' \\
\hline
Gold & ``question: What was Isaac Nicola's city of birth? context: Isaac Nicola Romero (1916 in Havana, Cuba) was a prominent Cuban guitarist. </s>'' \\
\hline
\end{tabular}
\label{input_formats}
\end{table*}

\subsection{Datasets}
We use three datasets to test our weakly-supervised questions in the ZRE setting.
We first use the RE-QA dataset released by~\citet{levy-etal-2017-zero} for tail entity (TE) generation using QA models.  RE-QA provides ten folds with 84 relation types in the train, 12 on the dev, and 24 on the test splits. There is no overlap among relation types of these splits per fold. Each fold contains 840k sentences in the train, 6k sentences in the dev, and 12k in the test split. Half of the sentences are negative examples with no corresponding $e_2$ given $e_1$ and $r$. The RE-QA dataset provides multiple gold question templates for each relation type.

To compare our system with current SOTA methods for the ZRE, we also consider the FewRel and WikiZSL datasets. The FewRel dataset has 56K sentences over 80 relation types~\cite{han-etal-2018-fewrel}. The WikiZSL dataset has 93483 sentences over 113 relation types~\cite{zhongSeq2SQL2017}. Similar to the previous SOTA method ZS-BERT~\cite{chen-li-2021-zs}, we randomly split the FewRel and WikiZSL datasets into train/dev/test splits where we have 5 and 15 relation types in the dev and test splits, respectively; with no overlap of the types between these splits. We create five train/dev/test folds splitting the datasets with different random seeds.

For all the datasets, if possible, we retrieve the relation description from the wikidata\footnote{\url{https://www.wikidata.org/wiki/Wikidata:List_of_properties/all_in_one_table}} and append it to the relation keyword $r$ in the question generator $P_{\theta_Q}$ or the sampling module $S(q)$. 

Table \ref{q_input_formats} lists an example for the input given to our question and search modules. Table \ref{input_formats} also provides an example for the input of the answer module for the gold, pseudo or generated questions.

\subsection{Evaluation Metrics}
For the TE task, as we generate a single tail entity $e^{'}_2$ for each input sentence, the extracted triple $(e_1, r, e^{'}_2)$ matches the ground truth triple $(e_1, r, e_2)$ only if $e^{'}_2 = e_2$ (case insensitive). For the negative examples, we generate a special `NO\_ANSWER' output specifying the null tail entity. We then use the official evaluation script\footnote{\url{http://nlp.cs.washington.edu/zeroshot/}} to compute the Precision, Recall, and F1-score for the TE task when there are negative examples in the test data. Precision is the
true positive count divided by the number of times
the system generates a non-null tail entity. Recall is
the true positive count divided by the number of
positive examples having the tail entity given $e_1$ and $r$. For the ZRE performed by the inference objective \ref{re-class-inference}, we compute the macro Precision, Recall, and F1-score; averaged across relation types only on positive examples.

\subsection{Baselines}
Few related works have focused on ZRE. The ZS-BERT model \cite{chen-li-2021-zs} maps sentences and relations into a shared semantic space using two BERT encoders and then uses
nearest neighbor search to predict the unseen relations. During training, the ZS-BERT minimizes the distance between the attribute vectors of the input sentence and the seen ground-truth relation while maximizing the distance from other incorrect relation representations. The ZS-BERT has outperformed another earlier work that treats the task of ZRE as an entailment task~\cite{obamuyide-vlachos-2018-zero}.

We also provide the reported ZRE results on the FewRel and WikiZSL datasets from \citet{tran-etal-2022-improving} which introduces discriminative inter-relation and inter-sentence losses along with a comparative network to better separate sentence-relation representation pairs.

We finally consider RelationPrompt~\cite{chia-etal-2022-relationprompt} which is trained to generate the relation triplets for a given sentence using a BART-base sequence-to-sequence model. We also apply its data-generation variant which uses GPT2 to augment the training dataset by generating synthetic (sentence, entity) pairs for the unseen test relations.

\begin{table*}
\centering
\caption{Example generated questions for the relationship type ``collection'' (``art, museum or bibliographic collection the subject is part of'') given the head entity ``The Vision of Saint Eustace.'' The semantically correct questions are in blue. The Base-Base model predicts the test data using only the pre-trained question-answer modules. The model name ``A-B'' denotes questions generated by the question module trained with the objective A while the answer module is trained with the objective B.}
\begin{tabular}{ p{0.22\linewidth} | p{0.70\linewidth} }
\hline
Q-A Models & Generated Question\\
\hline
Base-Base& ``What collection is part of The Vision of Saint Eustace?''\\
MML-MML& ``What subjects subjects?''\\
MML-G& ``What art collection is The Vision of Saint Eustace?'' \\
\hline
OffMML-OffMML& ``\textcolor{blue}{The Vision of Saint Eustace is part of what collection?}'' \\
OffMML-G& ``\textcolor{blue}{The Vision of Saint Eustace is part of what collection?}'' \\
GoldQ& ``\textcolor{blue}{What is the name of the place where The Vision of Saint Eustace can be found?}'' \\
\hline
\end{tabular}
\label{off-on-examples}
\end{table*}

\begin{table}[t]
\centering
\caption{The average Precision, Recall and F1-score for the tail entity generation task using our four training objectives on the ten test folds of the RE-QA dataset. The last column (PP) reports the average perplexity of the generated questions over these ten test folds. Best performance bolded, second best underlined.}
\begin{tabular}{ l | c c c | c}
\hline
Q-A Models & P & R & F1 & PP\\
\hline
Base-Base & 24.4 & 31.4 & 27.5 & 170 \\
MML-MML & 57.3 & 52.3 & 54.6 & 10456 \\
MML-G & 57.0 & \underline{52.9} & 54.7 & 1309 \\
OffMML-OffMML & \bf{60.7} & 52.3 & \underline{55.9} & \underline{148} \\
OffMML-G & \underline{58.3} & \bf{54.4} & \bf{56.2} & \bf{144} \\
\end{tabular}
\label{train-comparisons}
\end{table}

\subsection{Training Details}
For pre-training/fine-tuning T5 on QA and RE datasets, we use the Adafactor optimizer, which requires less memory compared to Adam~\cite{pmlr-v80-shazeer18a}. In addition, we follow the original configuration\footnote{\url{https://discuss.huggingface.co/t/t5-finetuning-tips/684/3}} suggested for fine-tuning the T5 transformers, such as disabling update clipping, no weight decay, and no warm-up starts~\cite{JMLR:v21:20-074}. In all of our experiments, we used the fixed learning rate of $0.0005$.

For fine-tuning our models and all the baselines on all the RE datasets, we train them for one epoch using the batch size of 16 on the RE-QA, 4 on the FewRel, and 16 on the WikiZSL datasets, respectively. Every 100 training steps, we evaluate the models on the dev sets, and then we report the performance of the dev set's best model on the test split.  For our models, on all the three datasets, we pre-train/fine-tune T5-small with 6 transformer blocks and 512 hidden states. Due to GPU constraints, we use eight samples or a beam size of 8 in the top-p sampling and beam search decoding algorithms, respectively.

\section{Results}
\subsection{Training Objective Comparisons}
In our first experiment, we compare the different training objectives listed in Table~\ref{objectives} in the tail entity generation task. To provide a fair comparison between these objectives, we also append the relation description to the relationship word $r$ in the on-policy training objectives MML-MML and MML-G. We compare these four objectives on the ten folds of the RE-QA dataset. We also report the baseline Base-Base that predicts the test data using only the pre-trained question-answer modules without fine-tuning the modules over the RE dataset.

Table~\ref{train-comparisons} summarizes the metrics on the tail entity generation task for these training objectives.  OffMML-G, which uses off-policy samples from the pre-trained question module, outperforms the rest of the training methods with the average F1-score of 56.2. Our proposed off-policy sampling technique has improved the performance of the tail entity generator in the model trained with OffMML-G by an average gain of 1.5 F1-score compared to the model trained with the MML-G objective. Furthermore, on the positive examples, the accuracy for generating the correct tail entity is higher when we take the best question from the question module $P_{\theta_Q}$ since the OffMML-G outperforms OffMML-OffMML by an average gain of 2.1 Recall points. To assess the grammar in the generated questions, we compute the average perplexity over the test folds using the GPT2-large~\cite{radford2019language} language model. The last column ($PP$) in Table ~\ref{train-comparisons} shows that both the off-policy sampling and the G gradient objective contribute to a lower perplexity compared to the on-policy sampling and the MML gradient objective as the OffMML-G achieves the lowest perplexity of 144. This observation verifies our hypothesis that off-policy sampling helps to avoid ungrammatical spurious questions.

In Table \ref{off-on-examples}, we also provide an example of the generated questions for the following input context: ``The Vision of Saint Eustace is a painting by the early Italian Renaissance master Pisanello, now in the National Gallery in London.'' Given the head entity ``The Vision of Saint Eustace,'' and the relationship type ``collection.'' We verify that the proposed off-policy sampling technique is necessary to create semantically and grammatically correct questions as the OffMML-G objective could successfully generate the question ``The Vision of Saint Eustace is part of what collection?''

\begin{table}[t]
\caption{The average Precision, Recall and F1-score on the 10 test folds of the RE-QA dataset for the tail entity generation task. Highest performance bolded, second highest underlined.}
\centering
\begin{tabular}{ l | c c c}
\hline
Q-A Models & P & R & F1\\
\hline
PseudoQ-Base & 1.7 & 3.2 & 2.3\\
Base-Base & 24.4 & 31.4 & 27.5\\
GoldQ-Base & 36.1 & \underline{55.1} & 43.6\\
\hline
\citet{levy-etal-2017-zero} & 43.6 & 36.5 & 39.6\\
PseudoQ-Trained & 57.9 & 50.2 & 53.6\\
OffMML-G & \underline{58.3} & 54.4 & \underline{56.2} \\
GoldQ-Trained & \bf{60.1} & \bf{57.9} & \bf{58.9} \\
\end{tabular}
\label{concat-gold-main}
\end{table}

\begin{table}[t]
\caption{The average Precision, Recall and F1-score on the ten test folds of the RE-QA dataset for the Zero-shot Relation Extraction (ZRE) task.}
\centering
\begin{tabular}{ c | c c c}
\hline
Models & P & R & F1 \\
\hline
ZS-BERT & 34.7 & 32.2 & 33.3 \\
PseudoQ-Trained & 63.3 & 60.5 & 61.8 \\
OffMML-G & 66.7 & 64.2 & 65.4 \\
GoldQ-Trained & \bf{67.3} & \bf{65.0} & \bf{66.1} \\

\end{tabular}
\label{re-concat-gold-main}
\end{table}

\begin{table*}
\centering
\caption{The average Precision, Recall and F1-score for ZRE on five random runs over the FewRel and WikiZSL datasets. There are 15 unseen relations in the test folds. Highest performance bolded, second highest underlined. The $\dagger$ is the reported performance of ZS-BERT \cite{chen-li-2021-zs}. The $\star$ is the reported performance of RelationPrompt without using synthetic examples \cite{chia-etal-2022-relationprompt}.}

\begin{tabular}{ c | c c c | c c c | c }
\hline

 & \multicolumn{3}{c|}{FewRel}  & \multicolumn{3}{c|}{WikiZSL} & \\
Model & P & R & F1 & P & R & F1 & Avg F1\\
\hline

ZS-BERT$\dagger$ & 35.5 & 38.2 & 36.8 & 34.1 & 34.4 & 34.3 & 35.6\\
ZS-BERT (our run) & 41.9 & 39.5 & 40.5 & 28.8 & 26.5 & 27.4 & 34.0 \\
\citet{tran-etal-2022-improving} & 44.0 & 39.1 & 41.4 &  38.4 & 36.0 & 37.2 & 39.3 \\
RelationPrompt$\star$ & {\bf 66.5} & 40.0 & 49.4 & 54.5 & 29.4 & 37.5 & 43.5 \\
RelationPrompt (our run) & 53.2 & 45.1 & 48.3 & 52.4 & 38.8 & 44.5 & 46.4 \\
RelationPrompt (our run) (+supp) & 59.7 &  {\bf 61.0} &  60.3 & 59.7 & {\bf 60.5} & 60.0 & 60.2 \\
\hline
PseudoQ-Trained & 30.8 & 32.4 & 31.6 & 33.1 & 34.6 & 33.8 & 32.7 \\
OffMML-G & 29.4 & 30.4 & 29.9 & 28.2 & 28.7 & 28.4 & 29.2 \\
PseudoQ-Trained (+negs) & 62.8 & 58.4 & \underline{60.5} & \underline{62.6} & 59.5 & \underline{61.0} & \underline{60.8} \\
OffMML-G (+negs) & \underline{63.7} & \underline{59.2} & \bf{61.3} & \bf{63.6} & \underline{60.4} & \bf{61.9} & {\bf 61.6} \\
\hline
\end{tabular}
\label{fewrl}
\end{table*}

\subsection{Baseline Comparisons} 
We now compare systems trained with the OffMML-G training objective with the two primary baselines of using pseudo or gold questions to fine-tune the answer generator. 

For the tail entity generation task, Table \ref{concat-gold-main} presents the average F1 scores on the ten test folds of the RE-QA dataset, considering both the positive and negative examples. Using the test data's gold templates, the pre-trained answer generator (GoldQ-Base) achieves an average F1-score of 43.6, and after fine-tuning the answer module, the baseline with gold questions achieves the highest average F1-score of 58.9, outperforming the previous BiDAF method~\cite{levy-etal-2017-zero}, which had an average F1-score of 39.6. With the OffMML-G training objective, we boost the performance of the pre-trained answer module from the average F1-score of 27.5 to 56.2. The OffMML-G training objective outperforms the baseline with pseudo questions by an average of 2.6 F1-score.

For the ZRE task, as listed in Table \ref{re-concat-gold-main}, the model trained with OffMML-G objective outperforms the PseudoQ baseline by an average F1 score of 3.6. This superior performance verifies that using semantically relevant questions improves the performance of QA models while fine-tuning them on the RE datasets. Moreover, with our generated questions, OffMML-G achieves the competitive F1 score of 65.4 compared to the F1 score of 66.1 achieved by the system using gold questions.

In our final experiments, we fine-tune the question-answer modules with the OffMML-G objective on the FewRel and WikiZSL datasets which have 15 unseen relations in the test folds. As the FewRel and WikiZSL datasets do not provide any negative examples, we create synthetic negative examples by repeating every train instance once with its ground-truth relation label replaced with another relation class from the train data. This data augmentation approach is simple enough to include with our models as we generate the null tail entity for negative examples. 
The recent discriminate approaches such as ZS-BERT and the method of \citet{tran-etal-2022-improving} use margin-based or contrastive learning to separate the ground-truth relation class from other negative relation labels. As presented in Table \ref{fewrl}, our transferred QA models trained with OffMML-G objective over the mix of positive and negative examples significantly outperforms the ZS-BERT and the recent baseline of \citet{tran-etal-2022-improving} with an average gain of 26.8 F1 points on both the FewRel and WikiZSL datasets. We do not change the architecture of the underlying transformers as apposed to the comparative network of \citet{tran-etal-2022-improving} trained over the BERT encoders.

We finally compare against RelationPrompt~\cite{chia-etal-2022-relationprompt} which directly generates the relation class in the decoder of the BART-base model given the input sentence and the prompt information about the head and tail entities. The OffMML-G (+negs) achieves an average gain of 15.2 F1 points on both the FewRel and WikiZSL datasets (Table \ref{fewrl})\footnote{As expected from the model input format, our negative examples do not improve RelationPrompt.}. The idea of using synthetic (sentence, entities) pairs generated by a large language model for the test relation classes improves the performance of RelationPrompt (see +supp row in Table \ref{fewrl}). Such data augmentation to use synthetic examples can also be helpful for our method to further guide the question generation for the test relation classes.

In Table \ref{small_analysis}, we provide some example questions generated for the WikiZSL test data using our question module trained with OffMML-G (+negs) objective.

\section{Related Works}

{\bf Weakly-Supervised Semantic Parsing}:
\noindent
The objective in Equation \ref{main-tail-objective} follows a similar search-and-learn framework used in the recent works of weakly-supervised program synthesis and semantic parsing \cite{guu-etal-2017-language, misra-etal-2018-policy, NEURIPS2018_f4e369c0, wang-etal-2019-learning, wang-etal-2021-learning}. For example, in the task of text-to-SQL generation, a set of possible SQL solutions for a given natural query are generated (searched) and then those solutions are validated with an SQL verifier. 
The program verifiers in these recent works provide a deterministic score/reward to update the search module. However, in our objective, the feedback for the question module is another stochastic answer generator that needs to be trained along with the search module.

{\bf Reading Comprehension for RE}:
\noindent
The joint extraction of entities and relations has also been reduced as a multi-hop QA task using pre-defined question templates for a few entity and relation types present in the ACE and the CoNLL04 datasets~\cite{li-etal-2019-entity}. Another recent work shows that asking diverse question templates could further boost the performance of the QA models for the joint entity-relation extraction~\cite{ijcai2020-546}. Furthermore, QA frameworks have been applied to re-score and verify the extracted relations of a separate relation extractor by forming simple pseudo-questions concatenating the head entity with the relationship words \cite{DBLP:journals/corr/abs-2104-02934}.

Several works have done event argument extraction through QA \cite{liu-etal-2020-event, du-cardie-2020-event}. \citeauthor{du-cardie-2020-event} uses predefined question templates for the events. The work of \citeauthor{liu-etal-2020-event} first picks a template query word for the topic of the event and includes it as the prefix for the question, and then translates a descriptive span around the event word into a descriptive question suffix by learning an unsupervised translation model. Although \citeauthor{liu-etal-2020-event} applies their approach only to the task of event extraction on short sentences from the ACE dataset, they manually provide a relevant question prefix for the relation type.

{\bf Question Generation}:
\noindent
The Question Generation (QG) research aims at generating natural questions given a document such that the answer modules can find the answers to these questions. Recent systems build end-to-end neural sequence-to-sequence models for QG~\cite{du-etal-2017-learning}. Moreover, the current research first extracts the key phrases which could be the answer to the question and then feeds it as an extra input signal while training the QG models~\cite{Wang_Wei_Fan_Liu_Huang_2019}. Multiple QA corpora have been combined to train a single question generator conditioned on multiple answer types~\cite{murakhovska2021mixqg}. 

Earlier work suggests dual training of the QG and answer-sentence selection tasks, however, they could marginally improve both of the QG and QA tasks~\cite{DBLP:journals/corr/TangDQZ17}. Closely related to our work, another recent study translates freebase triplets into natural questions to augment the QA corpora~\cite{serban-etal-2016-generating}. However, this triplet-to-question translation task does not use any context passages. The rule-based entity extraction and QG have also been explored for open information extraction on financial datasets~\cite{gupta2021zeroshot}. However, its rules are not comprehensive for generating questions for complex relations.

Apart from training the QG models on supervised data, recent work uses policy gradient reinforcement learning to a) Optimize rewards related to the fluency and answer-ability of the generated questions~\cite{yuan-etal-2017-machine}, b) Expose the model to its errors~\cite{DBLP:journals/corr/abs-1709-01058}, and c) Include paraphrase probability to properly compare the generated questions with the ground truth ones~\cite{zhang-bansal-2019-addressing}.

Despite the previous studies on QG, we treat questions as latent variables without having access to gold questions given the input context and the relation triplets.


\section{Conclusion}
\label{conclusion}
This work introduces the OffMML-G training objective to fine-tune the question and answer generators for RE. Our method generates semantically relevant questions for the answer module given the head entity and the relationship keywords. We demonstrated that with these weakly-supervised questions, one could fine-tune QA models on the RE corpora, achieving competitive results in detecting unseen relations. Our future direction would deploy the technique on document-level entity-relation extraction, further exploiting the inference capabilities of QA models.

\section*{Limitations}
A major limitation in our method is that we need to provide three distributions during training: a) $P(q | c, e_1, r)$ as the question generator, b) $P(e_2 | c, e_1, r, q)$ as the answer module, and c) $S(q)$ as our fixed search module over questions. This generative approach (i.e. generating the tail entity) for relation extraction requires more compute resources compared to a direct discriminate approach to learn $P(r | c, e_1, e_2)$, however, such a direct approach cannot be used to transfer QA models into the RE task.

We have used the T5-small models in all our experiments. Further gain can be achieved by switching into T5-large models, however, we leave those large-scale experiments for future work.

\section*{Ethics Statement}
Many language models show biases in their output due to the data used to train them \cite{liang2021towards}.  It is possible that these biases could affect the RE results that we present here (e.g., producing poor performance for certain kinds of relations, or for entities with names different from the training data).

\bibliography{acl}
\bibliographystyle{acl_natbib}

\appendix

\section{Appendix}
\label{sec:appendix}

\begin{table*}
\centering
\caption{The generated questions with OffMML-G (+negs) on a random example of the WikiZSL test split for the context: ``Jerome Ellsworth Couplin III ( born August 31 , 1991 ) is an American football safety for the Philadelphia Eagles of the National Football League ( NFL ) .'' The ground-truth relation triple is (Philadelphia Eagles, member of, National Football League)}
\begin{tabular}{ p{0.25\linewidth} | p{0.65\linewidth} }
\hline
Relation Class & Generated Question\\
\hline
1- contains the administrative territorial entity (P150) & ``What is the name of the administrative entity that contains the Philadelphia Eagles?''\\
\hline
2- student (P802) & ``What is the name of the notable student(s) of Philadelphia Eagles?''\\
\hline
3- located next to body of water (P206) & ``What body of water is located next to the Philadelphia Eagles?''\\
\hline
4- location (P276) & ``What is the location of the object, structure or event for the Philadelphia Eagles?''\\
\hline
5- drafted by (P647) & ``Which team was drafted by the Philadelphia Eagles?''\\
\hline
6- member of (P463) & ``What organization is a member of the Philadelphia Eagles?''\\
\hline
7- parent astronomical body (P397) & ``What is the name of the major astronomical body the Philadelphia Eagles belong to?''\\
\hline
8- author (P50) & ``Who is the main author of a written work for the Philadelphia Eagles?''\\
\hline
9- sport (P641) & ``What sport does Brandon Lee Graham play for the Philadelphia Eagles?''\\
\hline
10- field of this occupation (P425) & ``What field of occupation does Brandon Lee Graham represent for the Philadelphia Eagles?''\\
\hline
11- founder (P112) & ``Who is the founder or co-founder of the Philadelphia Eagles?''\\
\hline
12- work location (P937) & ``What is the name of the location where people or organizations were involved in work for the Philadelphia Eagles?''\\
\hline
13- from fictional universe (P1080) & ``What is the subject's fictional entity in the Philadelphia Eagles?'' \\
\hline
14- award received (P166) & ``What award does Brandon Lee Graham receive from the Philadelphia Eagles?''\\
\hline
15- located on astronomical body (P376) & ``What is the name of the astronomical body located on the Philadelphia Eagles?''\\
\hline
\end{tabular}
\label{small_analysis}
\end{table*}

\end{document}